# INTERNATIONAL JOURNAL OF SCIENTIFIC RESEARCH

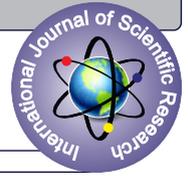

# EXTRACTION OF CUMULATIVE BLOBS FROM DYNAMIC GESTURES

| Computer Science | |
|---|---|
| **Rishabh Naulakha*** | Department of Information Technology, SRM Institute of Science & Technology. *Corresponding Author |
| **Shubham Gaur** | Department of Information Technology, SRM Institute of Science & Technology. |
| **Dhairya Lodha** | Department of Information Technology, SRM Institute of Science & Technology. |
| **Mehek Tulsyan** | Department of Information Technology, SRM Institute of Science & Technology. |
| **Utsav Kotecha** | Department of Information Technology, SRM Institute of Science & Technology. |

**ABSTRACT**

Gesture recognition is a perceptual user interface, which is based on CV technology that allows the computer to interpret human motions as commands, allowing users to communicate with a computer without the use of hands, thus making the mouse and keyboard superfluous. Gesture recognition's main weakness is a light condition because gesture control is based on computer vision, which heavily relies on cameras. These cameras are used to interpret gestures in 2D and 3D, so the extracted information can vary depending on the source of light. The limitation of the system cannot work in a dark environment. A simple night vision camera can be used as our camera for motion capture as they also blast out infrared light which is not visible to humans but can be clearly seen with a camera that has no infrared filter this majorly overcomes the limitation of systems which cannot work in a dark environment. So, the video stream from the camera is fed into a Raspberry Pi which has a Python program running OpenCV module which is used for detecting, isolating and tracking the path of dynamic gesture, then we use an algorithm of machine learning to recognize the pattern drawn and accordingly control the GPIOs of the raspberry pi to perform some activities.

## INTRODUCTION
Any Sufficiently Advanced Technology is Indistinguishable from Magic, if you think that that communication with computers is just doable by employing a keyboard, mouse, or touchscreen, you're fully wrong. Touchless technologies are simply round the corner and already penetrates our daily life: sensible cars, on-line client support, virtual assistants, sensible homes, increased and computer games. Signal acknowledgment is a perceptual UI, which depends on CV innovation that permits the PC to decipher human movements as orders, permitting clients to speak with a PC without the utilization of hands, therefore making the mouse and console pointless.

By differentiate voice acknowledgment, motion acknowledgment doesn't experience issues with recognizable proof. It can separate among individuals and activity. Motion acknowledgment's fundamental shortcoming is a light condition since motion control depends on PC vision, which intensely depends on cameras. These cameras are used to unravel flags in 2D and 3D, so the removed information can vary dependent upon the wellspring of light. The limitation of the structure can't work in a dull space. A basic night vision camera can be utilized as our camera for movement catch as they additionally impact out infrared light which isn't noticeable to people however can be unmistakably observed with a camera that has no infrared channel this significantly conquers the confinement of frameworks which can't work in a dull situation. In this way, the video stream from the camera is taken care of into a Raspberry Pi which has a Python program running OpenCV which is utilized for identifying, disconnecting and following the way of dynamic motion, at that point we utilize a calculation of AI to perceive the example drawn and in like manner control the GPIOs of the raspberry pi to play out certain exercises.

Hand motion has been the preeminent normal and regular methods for human to move and speak with each other, It gives communicator implies that of collaborations among those that includes hand stances and dynamic hand movements. A hand pose speaks to static finger arrangement while not hand development, though unique hand development comprises of a hand motion with or while not finger movement the adaptability to find and recognize the human hand signal show a few difficulties to specialists consistently. Signal acknowledgment is led with procedures from pc vision and picture process. In HCI, motion principally based interface offers a substitution heading towards the formation of a characteristic and easy to understand setting.

Gesture acknowledgment's fundamental shortcoming might be a lightweight condition because of motion the executives are predicated on PC vision, that intensely relies upon cameras.

These cameras are acclimated decipher motions in 2D and 3D, that the extricated data will change wagering on the stock of sunshine. The constraint of the framework can't include a dim environment.

## RELATED WORK
As opposed to the broad research on visual sensor-based hand signal acknowledgment, moderately barely any frameworks have been accounted for which can identify dynamic motions in a total dull condition. Currently, most technologies of the hand gesture recognition are supported standard cameras. However, RGB colour pictures from the standard camera have an obstacle that they cannot capture or recognise the gestures in low light or a dark environment, lot of the spatial position data needs to be inferred from 2D-3D mappings for the recognition throughout the imaging method due to which tons of data is lost. It is arduous to incorporate previous specific data in an exceedingly information driven learning calculation.

Dynamic hand signals are sketched out by structure varieties of the hand all through arrangements (for example fine motions performed by fingers), or by hand developments (for example swipe signals), anyway commonly each. These numerous qualities, that must be constrained to be thought about, form progressively solid the procedure of alternatives learning since it must be constrained to gain proficiency with each special and transient data. So as to thoroughly remove significant choices of cutting-edge hand signals utilizing information, models of neural system might want a larger than usual scope of layers which increment their time quality.

A few strategies present adequate runtime results utilizing profound systems yet utilize an amazing equipment with a few

GPUs. Right now, this equipment design is simply excessively sweeping and regularly impractical and, in this way, not fitting for genuine applications. We ought not overlook that, the client in front of the camera isn't constantly playing out a motion. Far more atrocious, the client performs also developments that don't appear to be pertinent and have a place with none of the motion classifications; for instance, to return back to a peaceful situation between important signals.

Therefore, the main motivating factor for creating a system which can detect and recognise patterns in dark environment and top of which it also overcomes all the limitations of
the current running systems

**LITERATURE REVIEW**
For realizing the problem statement, a total of four existing research paper were studied and observations were noted.

**a. Image process in Python with OpenCV**
This is a system for face detection and recognition using Open CV. It is used to detect and recognize human faces. The images of the persons are the datasets which are defined, trained and then detected. Face acknowledgment innovation has broadly stood out gratitude to its colossal application worth and market potential, similar to constant video reconnaissance frameworks. It's generally recognized that face acknowledgment includes a pivotal job inside reconnaissance frameworks since it doesn't care for the article's co-activity. It additionally shows that the change of hued picture to a grayscale picture is extremely straightforward.

The face recognition system is also being increasingly used in the mobiles for device security.

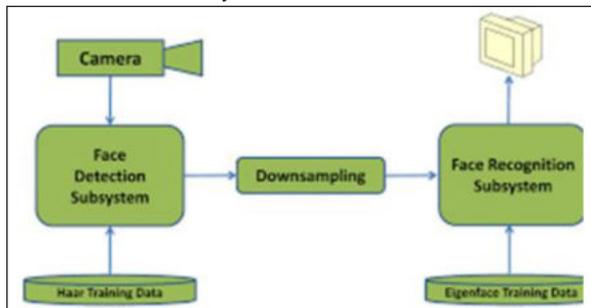

Pro: Face detected in live videos and images with high accuracy.

Con: The system is "lumination dependent" thus weaker in a dark environment**.**

**b. Use Hand Gesture to Write in Air Recognize with Computer Vision**
This proposed a framework which enables its user to literally write in the air through his/her figure and hand movement. The framework records strokes of the gestures to acknowledge written characters.

These characters may be single English alphabets, numeric or a whole connected word e.g., Hello Functionality of the framework begins with a user writing a personality in air, the Leap Motion Controller gets track the movements of the finger.

After pre-processing takes place frames are generated and feature extraction is done. Using computer vision techniques, the recognized text is then displayed on the screen.

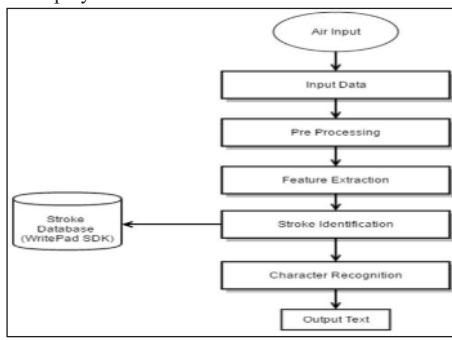

Pro: The evaluation of each alphabet separately and for most of the characters the system shows the average recognition rate is greater than 90%.

Con: The framework was getting a little confused in the recognition of some characters because of the resemblance in their shape.

**c. Detecting, Isolating and tracking line of action.**
This introduced an intuitive interface for expressive character posing based on the line of action metaphor. It gave an appropriate meaning of the line to furnish answers for some ambiguities related with this deliberation of the body; explicitly correspondence and profundity ambiguities. The methodology allows the client to quickly create communicatory postures beginning from exquisite style positions to misrepresented animation activities.

The line of action (LOA), an aesthetic curved stroke, allows the user to quickly specify the character's global shape in a single hand gesture. The pose can then be refined from other viewpoints and the remaining body parts can be posed using so-called secondary lines.

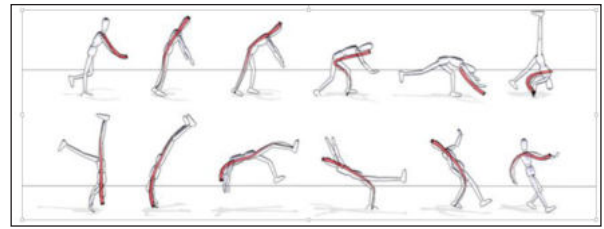

Pro: Optimization problem is that we can easily add additional constraints.

Con: A limitation of our viewing plane constraint is that bones can no longer twist when orthogonal to the viewing direction

| Serial No | Title Of the Paper | Concepts Proposed | Inference |
|---|---|---|---|
| 1. | Face Detection and Recognition Using OpenCV | Image Detection and Recognition | Image Processing using OpenCV |
| 2. | Use Hand Gesture to Write in Air Recognize with Computer Vision | Motion Detection | Video Based Motion Detection |
| 3. | The Line of Action: an Intuitive Interface for Expressive Character Posing | Detecting Line Of Action | Detecting, Isolating and tracking line of action. |

**PROBLEM STATEMENT**
The motion of a hand gesture is identified and converted into an image, by extraction of cumulative blobs from video frames, in the absence of light.

At a certain location, irrespective of its illumination, if a person performs a particular gesture with the sensor, the motion capture system should recognize the gesture and identify the pattern. All gestures correspond to a certain pattern which causes certain activities in the surrounding area like glowing of a led pane, opening of a box etc. activities.

To define the main objectives:
1. To extract images from line of action.
2. To solve the limitation of Computer Vision in lowlight (Infrared Technology).
3. Realizations into Home Automation, Driver Drowsy Alert and Detection, Human Computer Interaction for impaired people and children.

**PROPOSED SYSTEM**
The planned system is galvanized to counter the constraints of the prevailing system like static gesture recognition and its dependence on light-weight.

The planned system could be a dynamic gesture recognition system that uses infrared technology to eliminate the dependence on light-weight. Line of action is employed to make a pattern from the dynamic gesture retrieved from the live video frames.

The system has 2 elements: It can have a retroreflective bead and people retroreflective beads mirror an excellent quantity of actinic radiation that is given out by the second component i.e., the infrared camera within the motion capture system.

So, what we tend to understand as a not-so-distinctive device taking possession of the air, the motion capture system perceives as a bright blob which might be simply isolated within the video stream and half-track to acknowledge the pattern drawn by the person and execute the desired action. All this process takes place in real time and makes use of pc vision and machine learning. The planned system aims to realize a better accuracy than the prevailing system.

## IMPLEMENTATION METHODOLOGY

The infrared light produced by the motion capture camera is reflected in a great amount by the retroreflective beads. So, what we humans cannot see as a distinctive tip of the wand moving in the air, the motion capture system detects as a bright white blob which can be easily taken out in the video stream and identified to recognize the pattern drawn by the person and hence use it in further application. This process takes place by using technologies like Computer Vision and Machine Learning and happens in real-time.

A camera which has night vision ability can be used as our camera for motion capture as they produce infrared light which a human cannot see but is clearly visible with a camera that has no infrared filter just like Raspberry Pi NOIR Camera. So, the video stream from the camera is fed into a Raspberry Pi which has an OpenCV module running in Python program which is used for identifying, segregating and tracking the wand tip. Then we use SVM (Simple Vector Machine) lgorithm of machine learning in Python Language to recognize the pattern drawn and accordingly control the GPIOs (General Input and Output) of the raspberry pi to perform the required application. In this case, the GPIOs control a LED light.

A model is trained a machine learning model based on the Support Vector Machine (SVM) algorithm using a Dataset of handwritten English alphabets. The Dataset is in the form of .csv file which has 785 columns and more than 300,000 rows where every row represents a 28 x 28 image and every column in that row represents the value of that pixel for that image with an additional column in the beginning which contains the label, a number from 0 to 25, each corresponding to an English letter. Through a simple python code, the data is segregated to get all the images for only the 2 letters (A and C) which we want for this project and hence trained the model for them.

```
# SVM classifier created
clf = SVC(kernel='linear')
print("")
print("...Training the Model...")
clf.fit(train_features, train_labels)
print("...Model Trained...")

labels_predicted = clf.predict(test_features)
accuracy = accuracy_score(test_labels, labels_predicted)

print("")
print("Accuracy of the model using SVM is:  ")
print(accuracy)

print("...Saving the trained model...")
joblib.dump(clf, "alphabet_classifier.xml", compress=3)
print("...Model Saved...")
```

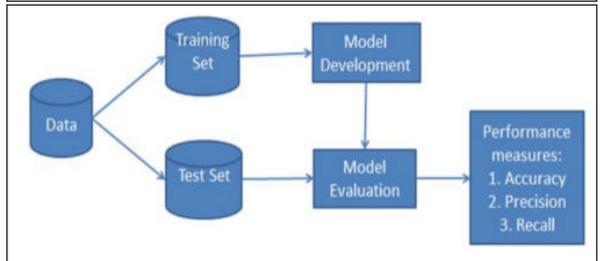

After creating the trained model with the help of Machine Learning Algorithms, the final step is to write a python program for our Raspberry Pi that helps us to do the following:

Access video from the picamera in real time Detect and track white blobs (which lights up in night vision) in the video Start detecting the path of the moving bright blobs in the video after some trigger event.

Stop detecting the path after the second trigger event.

Return the last frame which contains the pattern drawn by the user.

Perform the required pre-processing processes on the frame like thresholding, noise removal, resizing etc.

Use the processed last frame which contains the pattern drawn for prediction.

Perform some kind action (bulb lighten) by controlling the GPIOs of the Raspberry Pi according to the prediction.

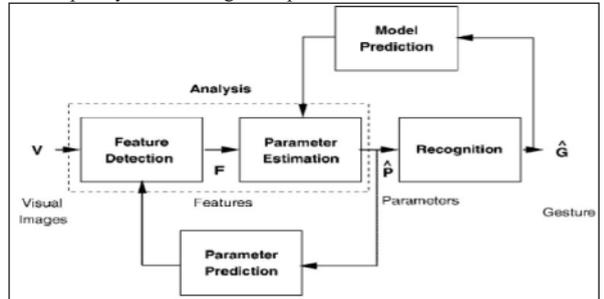

The trigger events which are mentioned above are created through two circles through python program in the real-time video where one circle is of green colour and the other one of red circle marking the starting and the ending points of the pattern. The green circle and the region within it mark the entrance of blob after which the user starts drawing the letter or pattern. After drawing the pattern when the blob approaches the red circle, the video stops and the last frame which contains the required pattern drawn by the user is passed to a function which performs the necessary pre-processing on the frame so that our frame is ready for prediction.

## RESULTS
The following results were obtained from the model built:

1. We built 3 modules, one was for training the dataset, the other was for configuration of camera and Raspberry PI, also for defining the trigger points in the camera window and the last one was for prediction and then realising into specified application.
2**.** The accuracy of the Machine Learning Algorithms which we used were, for Naive Bayes Classifier it was 64.20%, for Support Vector Machine Algorithm it was 99.62%, hence we chose Support Vector Machine Algorithm to train the model.
3. Users can make the desired pattern using any device with a retroreflective tip by starting the pattern at green trigger point and ending it at red trigger point.

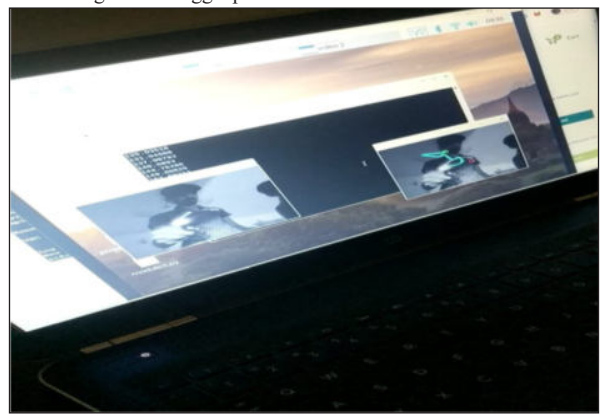

4. The pi camera has successfully been able to recognize blobs and form a single image of the video frames where the pattern would be saved in a number of JPEG files were the last frame JPEG file will contain the pattern.
5**.** This image file will be then sent to the prediction module where the final testing from the trained dataset will be done and if the pattern defined as 'A' the LED light will be switched on and if the pattern is described as 'C' the LED light will be switched off. The image pattern was successfully recognized by our machine learning algorithm.

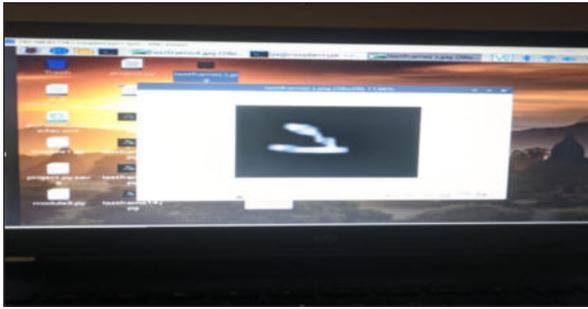

6. If we compare these results with the existing result, we find that there is no defined or prescribed way of getting the moving pattern from starting point into ending point in one single frame as the existing results either focused on the static gesture detection or detecting line of action. Through this model we were able to identify moving patterns into one image by extracting the cumulative blobs of dynamic gesture.

## CONCLUSIONS

The intend to empower the maximum capacity of PC vision and signal acknowledgment was accomplished with a high precision of 99.6% (Initially began with 64.2%). Another strategy for hand motion acknowledgment is presented in this paper.

Issues in existing frameworks were contemplated and a definite examination of it was likewise given. This framework was effective in tending to the previous issues. The signals are handled by NOIR-cameras that catch bars reflected from the retroreflective gadget. At that point, the line of activity is followed and the whole movement identified from the video outlines removes aggregate masses to shape a solitary picture. The acknowledgment of hand signals is practiced by a straightforward standard classifier. The exhibition of our strategy is assessed on a character informational index of pixels.

The AI calculations are right now prepared for just a couple of characters. The calculation determination for acknowledgment relies upon the application required. In this work application zones for the motion's framework are given. The exploratory outcomes show that our methodology performs well and is fit for the ongoing applications. Additionally, the proposed technique outflanks the current framework close by signals acknowledgment in obscurity.

This framework invalidates the need of light for the camera to catch motion and development.

The utilization of a retroreflective surface in any gadget to heighten and reflect IR light to the camera can work in complete obscurity. NOIR cameras give inside and out data that can improve the presentation of motion recognition.

## ACKNOWLEDGEMENT
The authors would wish to thank Dr G. Vadivu, Head of Department, Information Technology and Dr P. Supraja, Assistant Professor, Information Technology, SRM Institute of Science and Technology for their continuous support and guidance in our venture.